\documentclass[10pt,twocolumn,letterpaper]{article}

\usepackage{wacv}
\definecolor{wacvblue}{rgb}{0.21,0.49,0.74}
\usepackage[pagebackref,breaklinks,colorlinks,allcolors=wacvblue]{hyperref}

\usepackage{multirow}
\usepackage{makecell}
\usepackage{blindtext}
\usepackage{amssymb}
\usepackage{pifont}
\usepackage{fontawesome5}
\usepackage{rotating}
\usepackage[normalem]{ulem}

\newcommand{\mad}{\mbox{MVTec\hspace{1.5pt}AD}}
\newcommand{\madtwo}{\mbox{MVTec\hspace{1.5pt}AD\hspace{1.5pt}2}}
\newcommand{\loco}{\mbox{MVTec\hspace{1.5pt}LOCO\hspace{1.5pt}AD}}
\newcommand{\testpublic}{\mbox{\textit{TEST\textsubscript{pub}}}}
\newcommand{\testprivate}{\mbox{\textit{TEST\textsubscript{priv}}}}
\newcommand{\testprivatemixed}{\mbox{\textit{TEST\textsubscript{priv\_mix}}}}
\newcommand{\segfscore}{\mbox{$SegF_1$}}
\newcommand{\segfscoreprivate}{\mbox{$SegF_{1,\text{priv}}$}}
\newcommand{\segfscoreprivatemixed}{\mbox{$SegF_{1,\text{priv\_mix}}$}}
\newcommand{\xmark}{\ding{55}}
\newcommand{\cmark}{\ding{51}}
\newcommand{\vandchallenge}{\mbox{VAND\hspace{1.5pt}3.0~Challenge}}
\newcommand{\fonemax}{\mbox{$F_{1,max}$}}
\newcommand{\fonemaxavg}{\mbox{$\overline{F}_{1,max}$}}

\title{From Benchmarks to Reality:\\ Advancing Visual Anomaly Detection by the \vandchallenge}

\author{
Lars Heckler-Kram\textsuperscript{1,2},
Ashwin Vaidya\textsuperscript{3}, 
Jan-Hendrik Neudeck\textsuperscript{1},
\\
Ulla Scheler\textsuperscript{1},
Dick Ameln\thanks{work done at Intel}\hspace{4pt},
Samet Akcay\textsuperscript{3},
Paula Ramos\textsuperscript{4}
\vspace{0.25cm} \\
{
\textsuperscript{1}MVTec Software GmbH,
\textsuperscript{2}Technical University of Munich,
\textsuperscript{3}Intel,
\textsuperscript{4}Voxel51
}
}

\begin{document}
\maketitle

\begin{abstract}
Visual anomaly detection is a strongly application-driven field of research.
Consequently, the connection between academia and industry is of paramount importance.
In this regard, we present the \vandchallenge\ to showcase current progress in anomaly detection across different practical settings whilst addressing critical issues in the field.
The challenge hosted two tracks, fostering the development of anomaly detection methods robust against real-world distribution shifts (Category~1) and exploring the capabilities of Vision Language Models within the few-shot regime (Category~2), respectively.
The participants' solutions reached significant improvements over previous baselines by combining or adapting existing approaches and fusing them with novel pipelines.
While for both tracks the progress in large pre-trained vision (language) backbones played a pivotal role for the performance increase, scaling up anomaly detection methods more efficiently needs to be addressed by future research to meet real-time and computational constraints on-site.
\end{abstract}

\section{Introduction}
Visual Anomaly Detection serves as crucial tool for quality assurance in modern production systems.
By detecting deviations from the normal state of a product, it can be utilized for improving both product and process quality.
Thanks to the efforts of the scientific community, nowadays, a vast variety of deep learning-based anomaly detection methods exists and offers suitable solutions for many applications.
However, despite the ongoing research in this field, from a practical point of view limitations still exist that may hinder the deployment of an anomaly detection algorithm for certain real-world inspection scenarios.

\begin{figure}[t!]
    \centering
    \includegraphics[width=\linewidth]{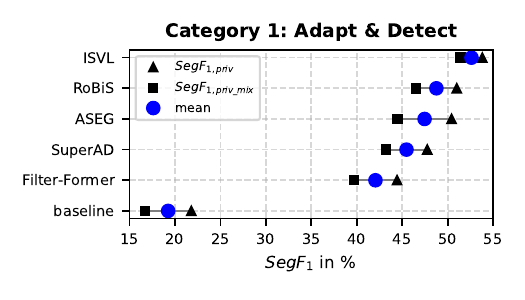}\\[-.5cm]
    \includegraphics[width=\linewidth]{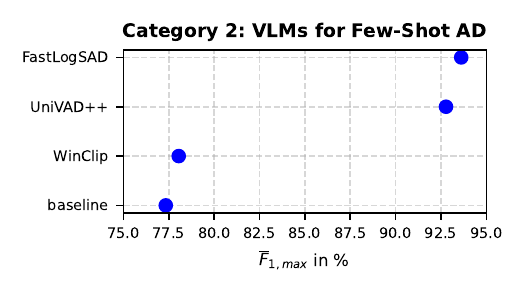}\\[-.5cm]
    \caption{Top submissions for Category 1 and Category 2 of the \vandchallenge. Both tracks resulted in Anomaly Detection methods with a significantly better performance than the baselines.}
    \label{fig:teaser}
\end{figure}

For this reason, we created the \textit{Visual Anomaly and Novelty Detection 2025 Challenge} (\vandchallenge) as part of the CVPR 2025 workshop on Visual Anomaly and Novelty Detection.
Contributing to bridging the gap between industry and academia, we hosted two tracks designed to foster the development of methods deployable on-site.
In particular, the design of the \vandchallenge\ focused on increasing the robustness against variations of the image acquisition settings (Category 1 — Adapt \& Detect: Robust Anomaly Detection in Real-World Application) as well as on detecting anomalies utilizing Vision Language Models (VLMs) while requiring only a small amount of training images (Category 2 — VLM Anomaly Challenge: Few-Shot Learning for Logical and Structural Detection).

Since in practice identifying all possible roots of failure and all possible defect states is hardly possible prior to launch of production, the \vandchallenge\ followed the setting of unsupervised anomaly detection, where exclusively images that do not contain any defects are used for training.
Besides, anomaly-free images are easier to collect and do not require cumbersome manual labeling.
At test time, models are not only required to classify images of products as \textit{good} or \textit{reject} but also to localize the defect precisely. 
A precise localization of a defect is crucial for enhancing the trust in the system's decision as well as to enable a systematic analysis of the manufacturing process via post-processing.

The \vandchallenge\ was open to practitioners and researchers from academia and industry alike and offered valuable insights in both the current state of visual anomaly detection algorithms for real-world inspection scenarios and problems that still need to be subject to future research.

Overall, the contribution of our work is three-fold:

\begin{itemize}
    \item Aiming towards bridging the gap between industrial and academic research, we organized and hosted the \vandchallenge\ as part of the CVPR 2025 workshop on Visual Anomaly and Novelty Detection.
    \item We explain the challenge design in detail and offer insights and transparency about the evaluation process which may serve as a guidance to other challenge organizers and opens the room for discussion about the organization of scientific technical challenges.
    \item  We discuss the challenge results with respect to the current state of anomaly detection algorithms from a practical point of view and outline promising directions for future research.
\end{itemize}

\section{Related Work}

\subsection{Unsupervised Anomaly Detection}
Similar to other fields in computer vision, progress in unsupervised anomaly detection (AD) is driven by the availability of suitable datasets.
Nearly perfectly solved today, \mad\ \cite{bergmann2019_mvtec_ad_cvpr} fostered the development of many AD approaches, before other datasets, e.g., VisA \cite{Zou_2022_VisA} or RealIAD \cite{wang_real-iad_2024} were curated and a broader range of problem settings, e.g. logical defects \cite{bergmann2021_mvtec_loco_ijcv}, 3D data \cite{bergmann2022_mvtec_3d_ad, bonfiglioli_eyecandies_2022}, or multiple views \cite{wang_real-iad_2024}, were considered.
Currently, new benchmarks such as \madtwo\ \cite{hecklerkram_mvtecad2_arxiv} or RobustAD \cite{RobustAD_Pemula_2025_CVPRW} try to stimulate new approaches for AD problems still not being sufficiently tackled by the scientific community with respect to practical deployment, e.g., threshold estimation or robustness against real-world distribution shifts.

Still, a large variety of AD algorithms exists.
They can be broadly categorized into memory bank-based \cite{patchcore, padim}, reconstruction-based \cite{Zavrtanik_DRAEM, Akcay_2019_skip-ganomaly, bergmann2018_ssim_ae} and distillation-based \cite{bergmann2020_uninformed_cvpr, deng2022_rd, luo2025INP-Former} approaches and all follow the same underlying paradigm: 
Since during training only anomaly-free data is seen, anomalous data evokes unusual patterns within the network which indicate a deviation from the distribution of normal data.
With the rise of Vision Language Models (VLMs) such as CLIP \cite{radford2021learningtransferablevisualmodels}, many recent AD methods incorporate these models to leverage the expressiveness gained by the large pre-training on these two modalities \cite{jeong2023_winclip}.

The \vandchallenge\ aims for exploring the boundaries of all these approaches with respect to real-world industrial inspection scenarios.
Based on state-of-the-art datasets, models are tested for robustness against distribution shifts (Category 1) and the applicability of VLMs in few-shot scenarios is examined (Category 2).

\subsection{Challenges in Computer Vision}
Over the past decade, computer vision challenges have played a pivotal role in accelerating progress across a wide range of tasks.
By offering standardized datasets, rigorous evaluation protocols, and competitive benchmarks, these challenges foster innovation, enable fair comparisons, and highlight emerging research directions.
They also serve as collaborative platforms that bring together academia and industry, often resulting in open-source tools and shared best practices.
Importantly, challenges aim to simulate real-world conditions, pushing researchers to develop robust, scalable, and generalizable solutions. 
Below is a selection of influential challenges that have significantly shaped the field of computer vision:

\textbf{Middlebury Stereo Vision (2001-present)}
One of the earliest benchmarks for stereo correspondence algorithms, Middlebury\footnote{\url{vision.middlebury.edu/stereo/eval3/}} provided high-quality ground-truth disparity maps and an online evaluation platform.
Until today, it remains a foundational resource for state-of-the-art stereo vision research \cite{Wen_2025_Foundation_Stereo_CVPR}.

\textbf{Pascal VOC Challenge (2005–2012)}
Pascal Visual Object Classes (VOC) Challenge \cite{everingham2010pascal} is a pioneering benchmark for object classification, detection, and segmentation.
It introduced standardized evaluation protocols and helped establish best practices that influenced later challenges.

\textbf{ILSVRC Challenge (2010–2017)}
The \mbox{ImageNet} Large Scale Visual Recognition Challenge (ILSVRC) \cite{russakovsky_imagenetChallenge_2015} revolutionized computer vision by introducing a massive labeled dataset for image classification and object detection - ImageNet \cite{imagenet_cvpr09}.
It was instrumental in the rise of deep learning, with landmark models like AlexNet \cite{AlexNet_NeurIPS2012} and ResNet \cite{he2016_resnet_paper} emerging from it.

\textbf{COCO  Challenge (2015–2020)}
The COCO (Common Objects in Context) dataset \cite{CoCo_Dataset_ECCV2014} advanced object detection and segmentation by introducing complex scenes with multiple objects and instance-level annotations.
Based on the dataset, versatile challenges were designed ranging from producing image captions over keypoint detection to dense estimation of human pose.\footnote{\url{cocodataset.org/}}

\textbf{BOP Challenge  (2017–present)}
Focused on estimating the 6D pose of rigid objects from RGB or RGB-D images, BOP (Benchmark for 6D Object Pose Estimation) \cite{nguyen2024bop_cvprw25} supports both model-based and model-free approaches.
Containing multiple industry-relevant datasets, it has become a key benchmark for 6D pose estimation with multiple tracks reflecting different real-world requirements for method development\footnote{\url{bop.felk.cvut.cz/}}.
\\
\\
Inspired by these significant contributions to scientific progress, the VAND Challenge has become an annual venue for benchmarking and developing models for visual anomaly detection based on real-world industrial inspection scenarios.

\subsection{Previous VAND Challenges}
The \vandchallenge\ is the third in a row of successful challenges on visual anomaly detection.
In its first edition in 2023\footnote{\url{sites.google.com/view/vand-cvpr23/challenge}}, the challenge emphasized building models that could generalize to unseen categories with little or no training data.
It featured two tracks: a zero-shot track, where models relied solely on textual descriptions without any training images, and a few-shot track, where models were trained using only 1, 5, or 10 normal images per category.
Participants were required to develop unified models capable of both anomaly classification and segmentation.

In 2024, the \mbox{VAND\hspace{1.5pt}2.0~Challenge}\footnote{\url{sites.google.com/view/vand-2-0-cvpr-2024/challenge}} featured two categories as well. 
Category 1 (Adapt \& Detect) challenged participants to build robust anomaly detection models that can handle real-world variability such as lighting changes, camera angles, and noise.
These models were trained only on normal images and evaluated on artificially perturbed test sets using the \mad\ dataset \cite{bergmann2019_mvtec_ad_cvpr}, emphasizing adaptability and consistent performance across diverse conditions.
Category 2 (VLM Anomaly Challenge) explored the use of few-shot learning and vision-language models (VLMs) to detect both structural and logical anomalies in industrial products using the \loco\ dataset~\cite{bergmann2021_mvtec_loco_ijcv}. 
Models needed to generalize from as few as 1 to 8 normal images per category, without pre-training on the \loco\ dataset itself.

The \vandchallenge\ (2025) builds upon the track specifications from 2024.
However, in contrast to \mbox{VAND\hspace{1.5pt}2.0} where distribution shifts were induced synthetically to the test data, Category 1 of the \vandchallenge\ incorporates real-world lighting changes as well as significantly more difficult defects contained in the \madtwo\ dataset~\cite{hecklerkram_mvtecad2_arxiv} to test the robustness of models.
Category 2 keeps the settings from the previous year. 
Hence, the VAND challenge steadily evolves with current state of the art while still connecting to core problems from previous editions.

\section{\vandchallenge}

The \vandchallenge\ took place as part of the CVPR 2025 workshop on Visual Anomaly and Novelty Detection\footnote{\url{sites.google.com/view/vand30cvpr2025}}.
Despite promising results from previous years, there remains significant room for improvement in developing robust and generalizable anomaly detection models.
Here, the \vandchallenge\ addresses critical industrial needs for reliable anomaly detection under varying conditions and with limited data.
In the following, the settings and results of the two challenge tracks are discussed.
Organizational details and challenge statistics are outlined in the supplemental material.

\subsection{Challenge Categories}

\begin{figure*}[ht]
    \centering
    \includegraphics[width=\linewidth]{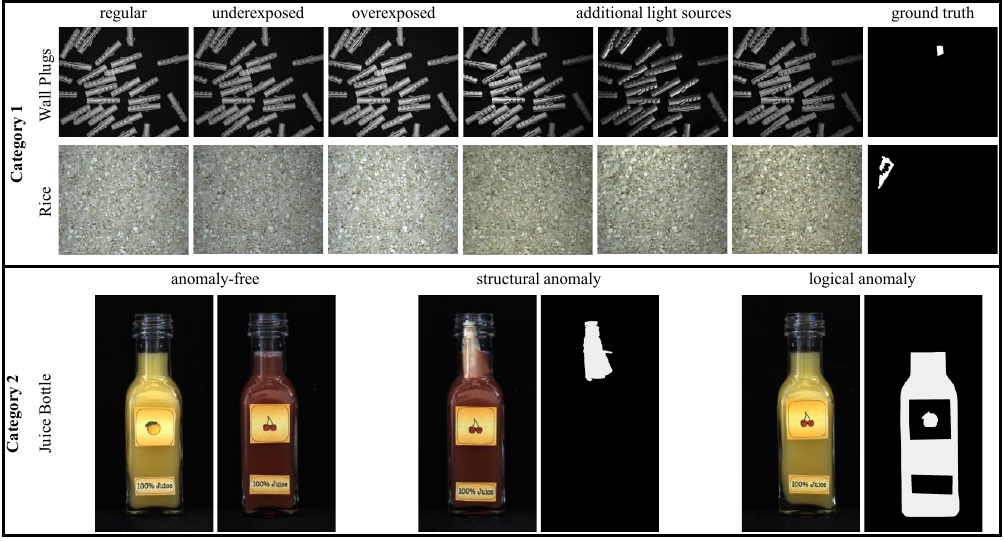}
    \caption{\vandchallenge\ Categories.
    \textit{top:} Category 1  utilizes the \madtwo\ dataset \cite{hecklerkram_mvtecad2_arxiv} to test models against real-world distribution shifts (Adapt \& Detect: Robust Anomaly Detection in Real-World Application).
    \textit{bottom:} Category 2 explores the applicability of Vision Language Models (VLMs) to detect structural and logical anomalies with a limited amount of training data based on the \loco\ dataset \cite{bergmann2021_mvtec_loco_ijcv} (VLM Anomaly Challenge: Few-Shot Learning for Logical and Structural Detection).
    }
    \label{fig:challenge_overview}
\end{figure*}

The two tracks of the \vandchallenge\ addressed different industry-relevant issues in practical anomaly detection (\cref{fig:challenge_overview}):

\begin{enumerate}
    \item Category 1 — Adapt \& Detect: Robust Anomaly Detection in Real-World Application
    \item Category 2 — VLM Anomaly Challenge: Few-Shot Learning for Logical and Structural Detection
\end{enumerate}

% Category 1 — Adapt \& Detect: Robust Anomaly Detection in Real-World Application
\noindent
Category 1 (\cref{sec:category_1}) focused on improving robustness of current anomaly detection models.
Although in contrast to open-world scenarios, the manufacturing process generally provides stable environmental conditions, certain changes in the acquisiton setting may still appear.
Apart from variations of the product pose within the field of view, the camera angle or focus might be subject to change \cite{hecklerkram_mvtecad2_arxiv}. 
Likewise, the aging of lights or spurious light sources potentially induce gradients in lighting over time.
Especially in industrial unsupervised anomaly detection, these domain shifts pose severe challenges since they might evoke false rejects in case models are not robust against such variations to the desired extend.

% Category 2 — VLM Anomaly Challenge: Few-Shot Learning for Logical and Structural Detection
Category 2 (\cref{sec:category_2}) targeted the development of models that are capable to not only cope with structural but also logical anomalies.
In contrast to structural anomalies that manifest themselves as a visible disruptions of known visual patterns logical anomalies violate underlying logical constraints, such as an interchanged position of certain components \cite{bergmann2021_mvtec_loco_ijcv}.
Here, Category 2 focused on leveraging vision language models to detect both types of anomalies.
Additionally, since even acquiring images of the normal state of a product requires significant monetary and time expenditure in certain cases, the amount of training data was strictly limited to assess the models' few-shot capabilities.

\section{Category 1: Adapt \& Detect}

\label{sec:category_1}
Category 1 aimed for the development of anomaly detection models that demonstrate robustness against external factors and adaptability to real-world variability.
Many existing anomaly detection models, trained on normal images and validated against normal and abnormal images, often struggle with robustness in real-world scenarios due to data drift caused by external changes such as camera angles, noise, or - in the focus of Category 1 - lighting conditions.

\begin{table*}[ht]
\centering
\caption{\madtwo\ dataset splits for each of the eight object categories. The private test sets (\testprivate, \testprivatemixed) do not provide public ground truth and evaluation is only possible via the submission portal. \testprivatemixed\ contains the same scenes as \testprivate\ but with lighting conditions randomly drawn from the 4-6 different conditions of each object category.}
\label{tab:mad_2_dataset_splits}
\begin{tabular}{lllc} 
\hline
dataset split        & description                             & lighting conditions & GT publicly available  \\ 
\hline\hline
train                & anomaly-free training images~           & regular                      & n.a.                     \\
validation           & anomaly-free validation images          & regular                      & n.a.                     \\
test public (\testpublic)        & public test set                         & regular, shifted\textsuperscript{*}            & \cmark\                   \\
test private (\testprivate)       & private test set w/o distribution shift & regular                      &  \xmark\                   \\
test private mixed (\testprivatemixed) & private test set w/ distribution shift  & regular, shifted\textsuperscript{*}            & \xmark\                    \\
\hline
\multicolumn{4}{l}{\small\textsuperscript{*}The number of induced lighting shifts varies per object category ($>=3$).} \\
\end{tabular}
\end{table*}

\subsection{Dataset}
\label{subsec:Cat1_dataset}
Category 1 built upon the \madtwo\ dataset \cite{hecklerkram_mvtecad2_arxiv}.
\madtwo\ is a public anomaly detection benchmark dataset that follows the design of previous popular anomaly detection datasets like \mad\ \cite{bergmann2019_mvtec_ad_cvpr} or VisA \cite{Zou_2022_VisA}.
In particular, it contains anomaly-free images for training and validation and both anomaly-free and anomalous images for testing. 

However, \madtwo\ aims to bridge academic research with industrial requirements in two ways.
First, it contains eight new challenging real-world scenarios captured under varying lighting conditions to reflect real-world distribution shifts.
Second, the ground truth of the official test set is non-public to emphasize the unsupervised nature of industrial anomaly detection, i.e., not knowing which defects to expect at inference time.
Only for development purposes, a small set of normal and anomalous test images with public ground truth is included in the dataset download.
\cref{tab:mad_2_dataset_splits} gives an overview about the design of the \madtwo\ dataset, which allows for assessing the robustness of a model against real-world lighting changes by comparing its performance on the private (\testprivate) and private mixed (\testprivatemixed) test set.
Images in \testprivate\ were acquired under the same lighting conditions as for the training images.
In contrast, \testprivatemixed\ contains images captured under both seen and unseen lighting conditions.

\subsection{Metrics}
\label{subsec:Cat1_metrics}

Model performance in Category 1 assessed the quality of anomaly localization and segmentation based on pixel level $F_1$ score (\segfscore) to ensure a balanced consideration of precision and recall:
\begin{equation}
    \segfscore = 2 \cdot \frac{\text{precision} \cdot \text{recall}}{\text{precision} + \text{recall}}
\end{equation}

It is noteworthy that \segfscore\ requires to select a threshold for the usually continuous anomaly maps – a challenge often not yet considered within the scientific community but indispensable for deployment in real-world applications.
Following standard evaluation protocols in anomaly detection, precision and recall were computed over the complete set of pixels in the respective test set and not averaged over individual images.

Ultimately, the final rank $\mathcal{R}_{\text{final}}$ of submissions in Category 1 of the \vandchallenge\ considered the overall performance as well as the robustness against real-world distribution shifts.
It was computed as the average rank of a model on \testprivate\ and \testprivatemixed\ in terms of the mean \segfscore\ over all eight object categories $c$ of \madtwo:
\begin{equation}
    \mathcal{R}_{\text{final}} = \frac{1}{2} [\mathcal{R}(\segfscoreprivate) + \mathcal{R}(\segfscoreprivatemixed)]
\end{equation}
with
\begin{equation}
    SegF_{1,t} = \frac{1}{8}\sum_{c = 1}^8{SegF_{1,c}}(\text{\textit{TEST}}_t)
\end{equation}
and $t$ specifying the test split (\textit{priv}, \textit{priv\_mix}) according to \cref{tab:mad_2_dataset_splits}.
In case of equal $\mathcal{R}_{final}$ the smaller absolute difference of the two \segfscore\ scores on \testprivate\ and \testprivatemixed\ determined the leaderboard position.

\subsection{Submission Platform}
\label{subsec:Cat1_submission_platform}
The official benchmark server\footnote{\url{benchmark.mvtec.com}} of \madtwo\ served as the submission platform of Category 1 of the \vandchallenge.
Here, participants were required to upload both the predicted continuous and thresholded anomaly maps for the two test sets \testprivate\ and \testprivatemixed.
Consequently, the thresholded anomaly images were compared with the non-public segmentation ground truth and evaluation metrics were computed in accordance with \cref{subsec:Cat1_metrics}.
A leaderboard entry was created automatically with the option to modify (except for the model predictions) or delete the entry again.

To highlight the concept of unsupervised anomaly detection, i.e., not knowing which defects and test data to expect, the evaluation budget per account was limited.
Participant were only allowed to make 2 submissions per week, which reduced the possibilities for extensive hyperparameter tuning on the official test data of \madtwo.

\subsection{Evaluation Protocol}
\label{subsec:Cat1_evaluation_protocol}
Models submitted to Category 1 were required to follow the one-class training paradigm, i.e., training exclusively on normal images, no further restrictions applied.
First, model performance was ranked according to the metrics described in \cref{subsec:Cat1_metrics}.
Second, submissions were checked for completeness, i.e., providing open-source code and a technical report.
Third, the reproducibility and technical soundness of the submitted material was verified in a review process.
Submissions adhering to all criteria were considered as valid submissions.

\subsection{Results}
\label{subsec:Cat1_results}
\begin{table*}[tb]
\renewcommand{\arraystretch}{1.0}
\centering
\caption{Official final leaderboard of \vandchallenge\ Category 1, only valid submissions are shown. Inspired by \cite{RobustAD_Pemula_2025_CVPRW}, $|\Delta_\text{rel}|$ is the relative difference between the performance without any changes in the environmental conditions (\segfscoreprivate) and performance on the test set with varied lighting conditions (\segfscoreprivatemixed). Baselines in the lower block are taken from \cite{hecklerkram_mvtecad2_arxiv}. Best per column is marked \textbf{bold}, second best is \uline{underlined}, best baseline is indicated by *, respectively.}
\label{tab:Cat1_results}
{
\small
\begin{tabular}{c|c|c|c|c|c}
\hline
\rule{0pt}{10pt} $\mathcal{R}_{\text{final}}$ & Method & \segfscoreprivate & \segfscoreprivatemixed & $|\Delta_\text{rel}|$ & Code \\
\hline
\hline
1 & ISVL \cite{Cat1_first_ISVL} & \textbf{53.81} & \textbf{51.43} & \textbf{4.42\%} & \href{https://www.github.com/ISVL119/isvl}{\scriptsize\texttt{</>}} \\
2 & RoBiS \cite{Cat1_second_RoBiS} & \uline{51.00} & \uline{46.52} & 8.78\% & \href{https://www.github.com/xrli-U/RoBiS}{\scriptsize\texttt{</>}} \\
3 & ASEG \cite{Cat1_third_ASEG} & 50.43 & 44.49 & 11.78\% & \href{https://www.github.com/vghost2008/ASEG}{\scriptsize\texttt{</>}} \\
4 & SuperAD \cite{Cat1_fourth_SuperAD} & 47.75 & 43.19 & 9.55\% & \href{https://www.github.com/Summerdayhurricane/SuperAD}{\scriptsize\texttt{</>}} \\
5 & Filter-Former \cite{Cat1_fifth_filter_former} & 44.43 & 39.68 & 10.69\% & \href{https://www.github.com/jcjing-commit/Filter-Former}{\scriptsize\texttt{</>}} \\
\hline
\hline
- & PatchCore \cite{patchcore} & 3.7 & 3.4 & 8.11\% & \href{https://www.github.com/amazon-science/patchcore-inspection}{\scriptsize\texttt{</>}} \\
- & RD \cite{deng2022_rd}& 18.1 & 16.7* & \uline{7.73\%}* & \href{https://www.github.com/hq-deng/RD4AD}{\scriptsize\texttt{</>}} \\
- & RD++ \cite{tien2023_rd_revisited} & 19.2 & 16.7* & 13.02\% & \href{https://www.github.com/tientrandinh/Revisiting-Reverse-Distillation}{\scriptsize\texttt{</>}} \\
- & EfficientAD \cite{batzner2023_efficientad}& 15.4 & 8.0 & 48.05\% & - \\
- & MSFlow \cite{zhou2024_msflow} & 21.8* & 9.0 & 58.72\% & \href{https://www.github.com/cool-xuan/msflow}{\scriptsize\texttt{</>}} \\
- & SimpleNet \cite{Liu_2023_simplenet} & 17.7 & 8.7 & 50.85\% & \href{https://www.github.com/DonaldRR/SimpleNet}{\scriptsize\texttt{</>}} \\
- & DSR \cite{Zavrtanik_2022_DSR} & 10.9 & 9.5 & 12.84\% & \href{https://www.github.com/VitjanZ/DSR_anomaly_detection}{\scriptsize\texttt{</>}} \\
\hline
\end{tabular}
}
\end{table*}
\cref{tab:Cat1_results} summarizes the results of Category 1 of the \vandchallenge.
All valid submissions exceeded the baselines by far with an overall best \segfscoreprivate\ of 53.81\% and an overall best \segfscoreprivatemixed\ of 51.43\%, highlighting the contribution of the \vandchallenge\ to anomaly detection research.
In the following, a short description of each methodology is provided, ordered by their appearance in the final leaderboard.

% \vspace{1em}
\textbf{1. ISVL} \cite{Cat1_first_ISVL} extends CPR \cite{cpr} and INP-Former \cite{luo2025INP-Former}, achieving superior segmentation performance through a tiling strategy to handle high-resolution images and an innovative morphological post-processing of the derived segmentation maps.

\textbf{2. RoBiS} \cite{Cat1_second_RoBiS} leverages INP-Former \cite{luo2025INP-Former} as a baseline, enhancing robustness to lighting variations through advanced data augmentation techniques.
Additionally, it applies a novel method for thresholding anomaly maps by combining classical blob analysis with deep learning-based refinement.

\textbf{3. ASEG} \cite{Cat1_third_ASEG} presents an ensemble approach combining GLASS \cite{chen2024glass} and INP-Former \cite{luo2025INP-Former}.
By integrating stronger transformer-based backbones, improved feature fusion via multi-layer convolutional networks, and refined training strategies, detection accuracy and robustness is enhanced.

\textbf{4. SuperAD} \cite{Cat1_fourth_SuperAD} is a training-free approach, leveraging the powerful feature extraction capabilities of the DINOv2 model \cite{oquab2024dinov2learningrobustvisual} and a memory bank built from diverse normal reference images inspired by \mbox{PatchCore \cite{patchcore}}.

\textbf{5. Filter-Former} \cite{Cat1_fifth_filter_former} builds upon INP-Former \cite{luo2025INP-Former}.
Trained with extensive data augmentation, a Diff Predictor Module additionally performs self-attention on encoder outputs to highlight discrepancies between normal and test images.

\subsection{Discussion}
\label{subsec:Cat1_discussion}
In Category 1 of the \vandchallenge, the results show significant improvements over baseline methods in threshold-dependent performance evaluation (\segfscore) and robustness.
%data augmentation
The extensive usage of data augmentation techniques helped to diversify the distribution of normal data seen during training with respect to different environmental conditions, i.e., lighting scenarios.
Though the relative difference between performance on test data from seen and mainly unseen lighting conditions $|\Delta_\text{rel}|$ could be reduced, there is still room for improvements.
Exploring data augmentation techniques even further might enable to close the gap. 

Generally, anomaly segmentation performance measured as \segfscore\ more than doubled compared to baselines.
On the one hand, this highlights the technical quality of submissions.
New post-processing strategies for anomaly maps were examined and the benefits of considering established classical techniques were demonstrated as well. 
% large image sizes --> runtime and memory
On the other hand, certain components of the solutions need to be assessed critically and from multiple perspectives.
First, all solutions of Category 1 considered large image sizes ($>= 448 \times 448$ pixels), either as one input or by tiling the original image.
Certainly, this is helpful to detect some of the defects contained in \madtwo\ \cite{hecklerkram_mvtecad2_arxiv}, however, in combination with the strong pre-trained backbones used, the question arises whether such approaches are still employable under real-time constraints or when having only access to limited compute.
In this regard, no submission specified the runtime or memory consumption of their method.
% rather no methodological contributions, as expected
Second, potentially due to the limited amount of time, most solutions built upon existing promising state-of-the-art approaches.
From a practical point of view, solutions that are ready to be deployed in praxis were created.
Nevertheless, from a scientific point of view, it will be interesting to see novel methodological concepts for further increasing robustness and localization accuracy.
% hybrid approaches
Third, likewise, as a technical challenge no constraints except the one-class training paradigm were imposed on method development.
Here, some hybrid approaches chose to select different model architectures for different inspection scenarios, i.e., object categories in \madtwo.
While this technique is valid from a practitioner's perspective, it simultaneously highlights the challenge of unifying the strengths of multiple architectures within a single, versatile solution as an open research question.
\begin{table*}[ht]
\renewcommand{\arraystretch}{1.0}
\centering
\caption{Official final leaderboard of \vandchallenge\ Category 2. Only the top two submissions are shown. Baselines in the lower block are taken from \url{https://cvpr-vand.github.io/challenge/} and VAND 2.0 Challenge results (only average Area Under F-Score Curve available). Best per column is marked \textbf{bold}, second best is \uline{underlined}, best baseline is indicated by *, respectively.}
\label{tab:Cat2_results}
{
\small
\begin{tabular}{c|c|c|c|c|c|c|c|c}
\hline
%\faGithub\
\rule{0pt}{10pt} $\mathcal{R}_{\text{final}}$ & Method & $\overline{\text{AUFC}}$      & \fonemaxavg\ & $k=1$ & $k=2$ & $k=4$ & $k=8$ &  Code\\
\hline
\hline
1 &  FastLogSAD    &   \textbf{93.81}   & \textbf{93.61} & \textbf{92.88} & \textbf{93.30} & \textbf{93.72} & \textbf{94.55} & \href{https://www.github.com/cvpr-vand/challenge/pull/122}{\scriptsize\texttt{</>}} \\
2 &  UniVAD++ \cite{gu2025univadtrainingfreeunifiedmodel} & \uline{92.94} & \uline{92.77} & \uline{92.18} & \uline{92.59} & \uline{92.97} & \uline{93.36} & \href{https://github.com/cvpr-vand/challenge/pull/123}{\scriptsize\texttt{</>}} \\
\hline
\hline
- & AnomalyMoE & 81.8* & - & - & - & - & - & - \\ 
- & RJVoyagers & 79.6 & -  & - & - & - & - & - \\
- & Locore & 79.4 & - & - & - & - & - & - \\
- & MVTec LOCO Diffusion-AD & 78.6 & - & - & - & - & - & - \\
- & WinClip \cite{jeong2023_winclip}                        & 78.05                     & 78.04* & 77.97* & 78.12* & 78.03* & 78.06* & \href{https://www.github.com/cvpr-vand/challenge/pull/8}{\scriptsize\texttt{</>}} \\
- & Random Model                   & 77.33                        & 77.33 & 77.33 & 77.33 & 77.33 & 77.33 & \href{https://github.com/cvpr-vand/challenge/pull/25}{\scriptsize\texttt{</>}} \\

\hline
\end{tabular}
}
\end{table*}
\section{Category 2: VLMs for Logical Anomalies}
\label{sec:category_2}
Category 2 encouraged the development of novel methods to identify both structural and logical anomalies, e.g., having a bottle filled with orange juice labeled as cherry juice (\cref{fig:challenge_overview}).
Thus, detecting logical anomalies goes beyond mere pixel-level identification of anomalous patterns and involves a broader understanding of the context shown within the image.
In this regard, the advent of Vision Language Models (VLMs) has created new possibilities to tackle this task.
Given their extensive pre-training protocols, the capabilities of VLMs for detecting logical anomalies when only provided with a few anomaly-free samples for training is examined by Category 2.

\subsection{Dataset}
\label{subsec:Cat2_dataset}
Category 2 built upon the \loco\ dataset~\cite{bergmann2021_mvtec_loco_ijcv} that contains both structural and logical anomalies.
\loco\ is a public dataset featuring five distinct object categories including ground truth annotations for all test images.
Until today, it remains the only dataset allowing to benchmark models on the task of logical anomaly detection.

\subsection{Metrics}
\label{subsec:Cat2_metrics}
\fonemax\ score on image-level was used as evaluation metric in Category 2.
It provides a balanced measure of a model's performance by combining both precision and recall into a single metric - taking the maximum value over all possible thresholds $t$:

\begin{equation}
\label{equation:fonemax}
\fonemax = \max_t  2 \cdot \frac{\text{precision}_t \cdot \text{recall}_t}{\text{precision}_t + \text{recall}_t}
\end{equation}
The final rank $\mathcal{R}_{\text{final}}$ of submissions was based on the average \fonemax\ score (\fonemaxavg) computed across all five categories $C$, three seeds $S$, and all four k-shot settings $K$:
\begin{equation}
\label{equation:fonemaxavg}
\fonemaxavg =  \frac{1}{C\cdot S \cdot K} \sum_{c=1}^C \sum_{s=1}^S \sum_{k=1}^K \fonemax_{c,s,k}
\end{equation}

\subsection{Submission Platform}
\label{subsec:Cat2_submission_platform}
For Category 2 of VAND 3.0, we introduced a new submission system based on feedback from previous year's challenge.
In 2024, many submissions had their code redacted once the winners were announced, as the authors intended to publish their methods.
While they re-shared the code when contacted, the challenge guidelines made it explicit that the source code shall remain in the public domain. Additionally, evaluating submissions was a manual process that remained obscure to participants.

To address these issues, we introduced an Apache 2.0 Licensed GitHub repository\footnote{\url{github.com/cvpr-vand/challenge/}} containing the evaluation code. Participants were required to fork the project and create a pull request to the repository. This approach ensured that each submission remained publicly accessible and open to scrutiny, was reproducible on the participant's machine, and made all build failures immediately visible for correction.

An additional advantage of open-sourcing the evaluation platform also allows others to build on top and host their own challenge with minimal engineering efforts.

\subsection{Evaluation Protocol}
\label{subsec:Cat2_evaluation_protocol}
Category 2 tests the performance of the models during inference in k-shot settings.
The models are tested across all the categories of \loco\ with the k-shots being 1, 2, 4, and 8.
Additionally, the average is taken across three values of seeds (0, 42, 1234).
This gauges the consistency of the model performance over multiple runs. 
Besides, each entry was required to adhere to constraints balancing performance and runtime as expected in real-world industrial anomaly detection applications.
The models were limited to a single Nvidia RTX 3090 GPU, and the evaluation on all combinations of categories, seeds and k-shots needed to take less than five hours.
For Category 2, a technical report was not mandatory.

\subsection{Results}
\label{subsec:Cat2_results}
\cref{tab:Cat2_results} summarizes the results of Category 2 of the \vandchallenge. 
Compared to the results of 2024 and the baselines, model performances are significantly higher, with a maximum \fonemaxavg\ of 93.61\%.
While the second-place submission only differs from the winning entry by 0.84\%, it is consistently lower over the aggregate of the distinct few-shot cases.

All entries leveraged recent developments in VLM-based models like GroundingDINO \cite{liu2023grounding}, SAM \cite{kirillov2023segany} or CLIP \cite{radford2021learningtransferablevisualmodels} and included these architectures or a combination of them in their solutions.
Following is a short description of the winning entries:

\textbf{1. FastLogSAD} improves LogSAD \cite{zhang2025trainingfreeanomalydetectionvision} by incorporating BEiT \cite{bao2022beitbertpretrainingimage} features, introducing multi-feature projection, integrating zero-shot prior knowledge \cite{qu2025bayesianpromptflowlearning}, and a few optimizations in the post-processing.

\textbf{2. UniVAD++} \cite{gu2025univadtrainingfreeunifiedmodel} uses the Recognize Anything \cite{zhang2023recognize} to identify objects and then uses Grounded SAM \cite{ren2024grounded} models to generate masks of all the generated segments. On top of these results, they introduced Component-Aware Patch Matching and Graph-Enhanced Component Modeling to aggregate anomalies at different semantic levels to produce the final result.

\subsection{Discussion}
\label{subsec:Cat2_discussion}
% comparison to baseline results
The results of Category 2 show a significant increase over the baseline models. 
The top-performing submission achieved a 20\% higher image-level \fonemaxavg\ score than the highest-performing baseline, WinCLIP \cite{jeong2023_winclip}, which uses the CLIP \cite{radford2021learningtransferablevisualmodels} model to identify anomalous regions based on a prompt.
In contrast, both winning models incorporate versions of DINO and SAM in their architectures.
% high-level analysis
This highlights three observations: 
firstly, foundation models are well suited for few-shot anomaly detection tasks;
secondly, effective anomaly detection requires robust contextual understanding. Pre-conditioning models with prompts provides a rich task context;
and finally, integration of task-agnostic detection models like DINO enables creation of sophisticated task-chains, such as detection followed by localization. 
This is particularly relevant for industrial use-cases where objects are often less cleanly isolated than in research datasets.

% data conatmination in pre-training
Despite the benefits of VLMs for anomaly detection, unfortunate criticism is that data contamination cannot be verified, i.e., the \loco\ dataset may have been included in the training of these foundation models.
Thus, the efficacy of these models needs to be judged in specific industrial settings where the dataset is proprietary and disjoint from those used for training.

Nevertheless, the winning submissions represent a significant contribution to the area of logical anomaly detection.
Looking ahead, future iterations of the challenge should prioritize inference speed and computational requirements to better align with industry needs for edge computing. 
Moreover, since the location of the anomaly itself may not always be easily discernible, for logical anomaly detection, ways outside standard anomaly maps of providing a clear rationale for why a model flags an image as anomalous need to be explored.

\section{Conclusion}
We present the organization, evaluation and results of the \vandchallenge\ to contribute to bridging the gap between industry and academia.
Two tracks covered different real-world criteria for successfully deploying anomaly detection algorithms on-site.
Submissions to Category~1 showed both improved overall performance on challenging anomaly detection scenarios and increased robustness against real-world distribution shifts.
Participants of Category~2 provided enhanced models compared to previous years capable to detect structural and logical anomalies given only a limited amount of training data.
Though tremendous progress has been made by the scientific community, both tasks still remain far from being solved and offer possibilities for future research - ranging from more deeply exploring existing architectures over the creation of further suitable datasets to novel methodological approaches.
\subsubsection*{Limitations}
Submissions to both categories leveraged large pre-trained models, but \vandchallenge\ did not consider runtime or memory consumptions to assess the final rank of a method.
Future editions of the VAND Challenge will impose more hardware constraints or create incentives for developing efficient approaches.
Likewise, domains outside classical 2D anomaly detection may be explored, e.g., multi-view or 3D anomaly detection.
\subsubsection*{Learnings for Challenge Organizers and Participants}
Considering the organization of the \vandchallenge\ as a technical scientific challenge, same scientific standards applied as for regular publications.
Here, we see transparency and reproducibility as two key components.
However, from an organizer's perspective, it is neither feasible nor beneficial to exactly predefine the tolerated solution space for participants without harming creativity.
Consequently, we encourage challenge organizers to clearly highlight that when in doubt about what is admissible, participants are advised to reach out to the organizers before the deadline.
Vice versa, as long as submissions adhere to the general challenge guidelines, they need to be considered as valid, unless scientific integrity or common norms of the particular research domain are violated.

Additionally, the challenge timeline correlates with the versatility of submissions.
We observed that shorter timelines (6 weeks, see supplemental material) foster the exploration of existing methods for the given problem.
Allocating more time for participating likely results in more methodological progress.
Thus, organizers can utilize this trade-off to support the challenge's goals.

\section*{Acknowledgments}
% Workshop Organizers, participants, sponsors
We gratefully acknowledge the organizers of the VAND 2025 workshop for hosting an engaging and well-coordinated event that made this challenge possible.
We also thank all the participants for their valuable contributions and collaborative spirit, as well as the prize sponsors for their generous support and commitment to fostering innovation.

\appendix

\section*{Supplemental Material}

\section{Challenge Organization}
The \vandchallenge\ was organized and hosted by three different parties with versatile backgrounds and interests in the field of industrial anomaly detection.
Connected by our experience in practical anomaly detection research and method development, we collaboratively designed the challenge concept and served as the reviewers for the final submissions.

\subsection{Participation Requirements}
The \vandchallenge\ was open to both researchers and practitioners from academia and industry from all over the world without any region-specific restrictions.
Participants could choose to take part in a single category or enter both in two separate submissions, either individually or organized in teams.
To maximize the benefit of the \vandchallenge\ for the anomaly detection research community, participants were required to open-source their code alongside the submission of a technical report to enable reproducibility.

\subsection{Evaluation Criteria}
In general, submissions had to be reproducible in order to be eligible for the final leaderboard.
If judges could not reproduce the submission, the submission was disqualified.
Due to the distinct scenarios and setups, each category had separate evaluation metrics to evaluate model performance, as outlined in the main paper, respectively

\subsection{Communication Channels}
Though the submission platforms for the two categories were separated due to technical reasons, participants had the opportunity to register to the challenge via a centralized web-service to receive important information via e-mail.
Besides, a Discord channel was available to communicate with other participants and the organizers.
Additionally, participants were allowed to reach out to the individual organizers, e.g., via e-mail or LinkedIn.
Answers to such request were shared via the e-mail distributor or Discord in case they were of interest for the broader challenge audience.

\subsection{Timeline}
\cref{tab:challenge_timline} shows the timeline of the \vandchallenge.
For approximately 6 weeks participants could prepare their solutions before the organizers determined the final leaderboard of both categories based on the evaluation criteria. Official awards were handed over and winners could present their approaches as part of the workshop on Visual Anomaly and Novelty Detection at CVPR 2025\footnote{Solutions were presented as a pre-recorded video.}.

\begin{table}[ht]
\caption{\vandchallenge\ Timeline.}
\label{tab:challenge_timline}
\centering
\begin{tabular}{ll} 
\hline
time            & event                 \\ 
\hline\hline
April 7\textsuperscript{th} 2025 & Challenge Start       \\
May 26\textsuperscript{th} 2025  & Submission Deadline   \\
June 3\textsuperscript{rd} 2025  & Results Announcement  \\
June 12\textsuperscript{th} 2025 & Official Awards       \\
\hline
\end{tabular}
\end{table}

\subsection{Challenge Statistics}
\cref{tab:challenge_statistics} summarizes the number of teams, unique users and final leaderboard entries.
Due to distinct submission platforms for the two tracks, statistics are outlined per category.
Overall, the challenge received a lot of attention within the field of unsupervised anomaly detection, also indicated by over 120 in-person attendees at the corresponding workshop.

\begin{table}[ht]
\caption{\vandchallenge\ Statistics for both categories.}
\label{tab:challenge_statistics}
\centering
\begin{tabular}{lcc} 
\hline
                                   & Cat. 1 & Cat. 2  \\ 
\hline\hline
\# teams providing report and code & 10         &             28\textsuperscript{$\curlyvee$}\\
\# unique users                    & 42\textsuperscript{*}         &             10\\
\# final leaderboard entries       & 136\textsuperscript{\dag}        & 29          \\
\hline

\multicolumn{3}{l}{\small \makecell[l]{\textsuperscript{*}one team could create multiple accounts/ consist of\\ multiple users}} \\

\multicolumn{3}{l}{\small\makecell[l]{\textsuperscript{\dag}one account could submit multiple times and entries could be\\ deleted again}}\\

\multicolumn{3}{l}{\small\makecell[l]{\textsuperscript{$\curlyvee$}category 2 did not mandate report submission. Participants \\can submit multiple entries}}
\end{tabular}
\end{table}

\section{\vandchallenge\ Awards}
The \vandchallenge\ awards are based on the results presented in the main paper and first\textsuperscript{1} and second best\textsuperscript{2} approach were honored, respectively.
Further certificates of participation were issued upon request.

\textbf{Category 1: Adapt \& Detect}
Authors of the awarded entries: 
\textbf{ISVL}\textsuperscript{1} \cite{Cat1_first_ISVL} by Xingao Wang, Shuying Xia, Zhaohong Liao, Mengjie Xie, Handa Wang and Zhi Gao.
\mbox{\textbf{RoBiS}\textsuperscript{2}} \cite{Cat1_second_RoBiS} by Xurui Li, Zhongsheng Jiang, Tingxuan Ai and Yu Zhou.

\textbf{Category 2: VLM Anomaly Challenge}
Authors of the awarded entries: \textbf{FastLogSAD-v3.0c} by Xian Tao, Zhen Qu, Mengqi Song, Hengliang Luo, Dingrong Wang, Fei Shen, Zhengtao Zhang; \textbf{UniVAD++} \cite{gu2025univadtrainingfreeunifiedmodel} by Zhaopeng Gu, Bingke Zhu, Guibo Zhu, Yingying Chen, Ming Tang, Jinqiao Wang.

\clearpage
{
    \small
    \bibliographystyle{ieeenat_fullname}
    \bibliography{main}
}

\end{document}